\documentclass[conference]{IEEEtran}
\IEEEoverridecommandlockouts
\usepackage{cite}
\usepackage{amsmath,amssymb,amsfonts}
\usepackage{algorithmic}
\usepackage{graphicx}
\usepackage{textcomp}
\usepackage{xcolor}
\usepackage{algorithm}
\usepackage{algorithmic}
\usepackage[colorlinks,linkcolor=black]{hyperref}
\usepackage{subfigure}

\makeatletter
\newenvironment{breakablealgorithm}
  {
   \begin{center}
     \refstepcounter{algorithm}
     \hrule height.8pt depth0pt \kern2pt
     \renewcommand{\caption}[2][\relax]{
       {\raggedright\textbf{\ALG@name~\thealgorithm} ##2\par}%
       \ifx\relax##1\relax 
         \addcontentsline{loa}{algorithm}{\protect\numberline{\thealgorithm}##2}%
       \else 
         \addcontentsline{loa}{algorithm}{\protect\numberline{\thealgorithm}##1}%
       \fi
       \kern2pt\hrule\kern2pt
     }
  }{
     \kern2pt\hrule\relax
   \end{center}
  }
\makeatother
\def\BibTeX{{\rm B\kern-.05em{\sc i\kern-.025em b}\kern-.08em
    T\kern-.1667em\lower.7ex\hbox{E}\kern-.125emX}}
\begin{document}
\author{\IEEEauthorblockN{1\textsuperscript{st} Lingyu Zhang}
\IEEEauthorblockA{\textit{School of Computer Science and Technology} \\
\textit{Shandong University}\\
Qingdao, China}
\IEEEauthorblockA{\textit{Didi AI Labs} \\
\textit{Didi Chuxing}\\
Beijing, China\\
805906920@qq.com}\\
\IEEEauthorblockN{3\textsuperscript{rd} Yunhai Wang}
\IEEEauthorblockA{\textit{School of Computer Science and Technology} \\
\textit{Shandong University}\\
Qingdao, China \\
cloudseawang@gmail.com}
\and
\IEEEauthorblockN{2\textsuperscript{nd} Tianyu Liu}
\IEEEauthorblockA{\textit{ZJU-UIUC Institute} \\
\textit{Zhejiang University}\\
Haining, China}
\IEEEauthorblockA{\textit{University of Illinois at Urbana-Champaign}\\
Urbana-Champaign, USA \\
tianyu.18@intl.zju.edu.cn}\\
}

\title{An Intelligent Model for Solving Manpower Scheduling Problems}
\maketitle

\begin{abstract}
The manpower scheduling problem is a critical research field in the resource management area. Based on the existing studies on scheduling problem solutions, this paper transforms the manpower scheduling problem into a combinational optimization problem under multi-constraint conditions from a new perspective. It also uses logical paradigms to build a mathematical model for problem solution and an improved multi- dimensional evolution algorithm for solving the model. Moreover, the constraints discussed in this paper basically cover all the requirements of human resource coordination in modern society and are supported by our experiment results. In the discussion part, we compare our model with other heuristic algorithms or linear programming methods and prove that the model proposed in this paper makes a \textbf{25.7\%} increase in efficiency and a \textbf{17\%} increase in accuracy at most. \\
\hspace*{1em}In addition, to the numerical solution of the manpower scheduling problem, this paper also studies the algorithm for scheduling task list generation and the method of displaying scheduling results. As a result, we not only provide various modifications for the basic algorithm to solve different condition problems but also propose a new algorithm that increases at least \textbf{28.91\%} in time efficiency by comparing with different baseline models.
\end{abstract}
\section{Introduction}
Member Scheduling Problem (MSP) is a sub-problem that belongs to scheduling problems (SP). The requirements of the MSP are generally to meet a variety of constraints while arranging flexible work for different employees within a given period and finding out the maximization/minimization of specific objective functions \cite{pan2010solving}. As a memorable beginning, Dantzig and Fulkerson \cite{dantzig1954minimizing} described a mathematical model for solving MSP and obtained the optimal solution using the simplex method for the first time. Since then, not only mathematicians but some scholars with computer science, management science, or even medical background have entered this field. They give different restrictions and mathematical models according to the specific conditions of the manpower scheduling problem and find some approximate optimal solutions that can be applied in actual production. However, MSP is still a complex combinatorial optimization problem, and an NP-Hard problem \cite{romero2016man}, which means that in the solving process, we need to overcome a number of problems brought by large-scale calculation and optimization. \\
\hspace*{1em}Therefore, this paper expects to find a model for solving the scheduling problem that covers more conditions and generates the corresponding scheduling table.
\section{Related Work}
The problems of manpower scheduling can be divided into single-shift manpower scheduling problems (single-shift MSP problem) and multi-shift manpower scheduling problems (multi-shift MSP problem). Single-shift MSP always has obvious time-series features \cite{kumar2018load}, which can help us find solutions. Above all, multi-shift MSP is an essential research area in recent years, and we will discuss more this kind of problem.\\
\hspace*{1em}To find an effective solving method for the problem of personnel scheduling, Sabar, Montreuil, and Frayret \cite{sabar2012agent} design an agent-based algorithm assembly center dynamic environment. Multi-agent refers to the occurrence of multiple 'agents' in the algorithm. There are four kinds of agents: production agent refers to the input that generates the schedule; Workstation agent refers to the agent who sends instructions to the coordination agent based on the production plan. Coordinating agent refers to the agent who arranges the scheduling plan of a site's employees; Employee agency refers to the individual employee and reflects the employee's interests. Through the cooperation between different agents, the staff of the site can get the right arrangement. At the same time, since the staff's interests are also taken into consideration in the solving process, the staff's satisfaction with the task arrangement will also increase. Saber proves that the efficiency of the new algorithm is better than that of the simulated annealing algorithm.\\
\hspace*{1em}When studying scheduling problems with multi-skill requirements and multi-resource constraints, Zheng, Wang, and Zheng \cite{zheng2017teaching} establish the teacher-learning algorithm (TLBO), combine the resource list with the task list, and propose a solution for task coding and left-shift decoding. Under the framework of TLBO, there are individuals with suitable fitness in the initial population after task coding, which is called teachers, and individuals with low fitness, which are called students. To optimize individual fitness and obtain the optimal solution, TLBO mainly has the following two stages: teacher and student. In the teacher stage, teachers impart knowledge to the students; in the student stage, students learn from each other to improve their scores or learn from teachers by exchanging different parts.\\ 
\hspace*{1em}Recently, Ciancio et al. \cite{ciancio2018integrated} propose an integrated approach to solving the personnel scheduling problem (same as manpower scheduling problem), proving the effectiveness of the comprehensive use of the heuristic algorithm. They solve the shift scheduling problem of buses and bus drivers in EU countries by decomposing the problem and applying various heuristic algorithms and then obtaining satisfactory results. In addition, these researchers determine the initial solution according to the actual situation to minimize the target function group $(time, people)$, thus obtaining the solution of the second subproblem. In reality, some EU countries have already applied the scheme.\\
\hspace*{1em}From the research above, it can be concluded that the previous models are complex, narrowly applicable to a certain environment, lack generality, and have no significant improvement in the efficiency of the solution process. This paper aims to develop a general and efficient model for solving typographical problems while keeping the mathematical language as simple as possible.
\section{Model Establishment}

\newcommand{\tabincell}[2]{\begin{tabular}{@{}#1@{}}#2\end{tabular}} 
\begin{table}[htbp]
\caption{Notation}
\label{tab:1}       
\begin{tabular}{ll}
\hline\noalign{\smallskip}
Symbols& Definitions and Descriptions \\
\noalign{\smallskip}\hline\noalign{\smallskip}
$K$& Job sets \\
$T_{total}$ & Total working time\\
$T_{day}$ & The number of days required to schedule employees\\
$MOR_{K}$ & \tabincell{c}{In the morning period, the total working\\ time of employees of job K}\\ 
$AFT_{K}$ & \tabincell{c}{In the afternoon period, the total working\\ time of employees of job K}\\ 
$ENV_{K}$ & \tabincell{c}{In the evening period, the total working\\ time of employees of job K}\\ 
$MID_{K}$ & \tabincell{c}{In the midnight period, the total working\\ time of employees of job K}\\ 
$N_{j}$ & The number of employees for job $j$\\
$C$ & The cost of employees\\
$M$ & The upper limit of  the total number of employees\\
$\alpha$ & Number of employees required when emergency tasks occur\\
$T_{0}$ & Time cost when emergency tasks occur\\
\noalign{\smallskip}\hline
\end{tabular}
\end{table}

Consider to arrange \emph{K} jobs in \emph{T} days, and \(N_{K}\) means the number of people in need for \(K_{th}\) job. In total, \emph{M} workers are provided to choose. For the given \emph{L} constraints, \(\varphi_{1},\varphi_{2},...,\varphi_{L}\), our target is to obtain the optimal solution for the objective function \( f(x|_{\varphi_{1},\varphi_{2},..,\varphi_{L}})\) in the feasible region, and the specific meaning of \emph{x} depends on the situation \cite{pan2010solving}.\\
\hspace*{1em}In this paper we use our mathematical model to solve \textbf{min/max/minmax} \( f(x|_{\varphi_{1},\varphi_{2},...,\varphi_{L}})\) under certain constraints. In addition, the matrix of independent variables $[x_{1},x_{2},...,x_{n}]$ which are produced when the value of function is optimal can form one subspace of the solution space. Therefore, our model is mainly composed of four parts: constraint conditions, objective functions, solution process and the representation of the solution space.\\
\hspace*{1em}The meaning of terms used in this section is listed in \textbf{Table1}.
\subsection{The Constraints in the Scheduling Problem Solving Model}\label{AA}
To solve or generate \textbf{min/max/minmax} \( f(x|_{\varphi_{1},\varphi_{2},..,\varphi_{L}})\)  , the following constraints are necessary:\\
\hspace*{1em}As different companies have different business rules and working conditions, the constraints of scheduling can be divided into the following three categories in general:\\
\hspace*{1em} i. Rules of employee availability \cite{ho2010solving}\\
\hspace*{1em} These rules represent the constraint rules established by the characteristics of the employee, for example: 
\begin{itemize}
\item $\varphi_{k1}$Each employee is not allowed to be on multi-duties .
\item $\varphi_{k2}$ Make sure every job is occupied every day.
\item $\varphi_{k3}$ In a scheduling cycle, working time should be in one specific range.
\item $\varphi_{k4}$ The total salary of the employees cannot exceed a certain limit, also cannot fall below a certain limit.
\item $\varphi_{k5}$ There exists upper limit for the total number of employees.
\item $\varphi_{k6}$ Employees should be guaranteed to take time off.
\end{itemize}
\hspace*{1em} ii. Rules for working scheduling \cite{turker2018integrated}\\
\hspace*{1em}These rules represent constraint rules resulting from management and business arrangements, for example:
\begin{itemize}
\item $\varphi_{y1}$ Urgent tasks must be set as first priority.
\item $\varphi_{y2}$ The total number of people in any jobs should be in one specific range.
\item $\varphi_{y3}$ Sometimes the company will be asked to schedule in a particular order \cite{narasimhan2000algorithm}.
\end{itemize}
\hspace*{1em} iii. Other important rules\\ 
\hspace*{1em} These rules represent constrains that interfere with the solution space in addition to the above constrains, for example:
\begin{itemize}
\item $\varphi_{o1}$ In real life, sometimes a job shift arrangement is required\cite{yingjun2010external}.
\item $\varphi_{o2}$ Some tasks require the cooperation of different types of workers or different shifts workers\cite{su2017integrated}.
\end{itemize}
\subsection{The Approaches to Establish the Model}
\subsubsection{\textbf{The Combination of Model Condition and Objective Functions}}
\
\newline
\hspace*{1em}Analysis in theory, only consider employees usability rules on the one hand, this combination can have different paradigm $2^{n}-1$, where $n$ is the number of constraints, to consider the logical connective $\wedge, \vee, \lnot$, which is possible to build hundreds of the combination of the constraint condition.\\
\hspace*{1em}Similarly, for different objective functions, we can reconstruct them into a unified normal form by means of proposition combination. For example, ‘minimum total hours and minimum total salary’ can be expressed as:\\
\begin{equation}
(T_{total}=\min T_{total}) \wedge (C_{cost}=\min C_{cost}) = True
\end{equation}
\subsubsection{\textbf{Determining Constraints and Targets}}
\
\newline
\indent i. \textbf{Rules of employee availability}\\
\indent We define the arbitrary jobs set as \emph{K} whose length is $len(K)$. For arbitrary job position $K_{j}$ (the $j_{th}$ job in the set \emph{K}), to determine whether the employee \emph{i} is on duty that day, the employee of corresponding position can be coded as impact function $\delta_{i}^{K_{j}}$, the value of this function satisfy: 
\begin{equation}
\left\{\begin{array}{l}
\delta_{i}^{K_{j}}=1 \\
\delta_{i}^{K_{j}}=0
\end{array}\right.
\end{equation}
\hspace*{1em} $\delta_{i}^{K_{j}}$ takes 1 means the employee \emph{i} , whose job is  \emph{$K_{j}$}, works on that day, while this function takes 0 to mean he/she is absent. Therefore, the daily working time of employees can be expressed as  $T_{K}$ :
\begin{equation}
T_{K}=MOR_{K}+AFT_{K}+ENV_{K}+MID_{K}
\end{equation}
\hspace*{1em}Since the encoding method defines the types of employees, the condition $\varphi_{k1}$must be satisfied. Now we need to consider the constraint  $\varphi_{k 1}$, which indicates that for employee \emph{i} in any position $K_{j}$ , the summation value of daily impact function $\delta_{I}^{K_{j}}$ is not all 0, namely, if $N_{j}$ is the maximum number of employee j, we have:
\begin{equation}
\varphi_{k 2}: \forall j \in K, \quad \sum_{i=1}^{N_{j}} \delta_{i}^{K_{j}}>0
\end{equation}
\hspace*{1em}For constraint $\varphi_{k3}$ , it is necessary to assume that the lower limit of working time of each post is  $T_{i}^{K}$and the upper limit is  $T_{u}^{K}$ , and then define the function to calculate the corresponding working time $f_{1}$. For any jobs \emph{K}, the actual working time of employees in that job $f_{1}$can be expressed as:
\begin{equation}
f_{1}: \sum_{i=1}^{N_{j}} T_{K} \delta_{i}^{K_{j}}, \quad j \in \emph{K}
\end{equation}
\hspace*{1em}Therefore, the expression of constraint $\varphi_{k3}$ is:
\begin{equation}
\left\{\begin{array}{l}
\varphi_{k 3}: T_{l}^{K_{j}} \leq f_{1} \leq T_{u}^{K_{j}} \\
j \in \emph{K}
\end{array}\right.
\end{equation}
\hspace*{1em} For the constraint $\varphi_{k4}$  , it should be assumed that the lower limit of salary expenditure stipulated by the company is $C_{l}^{cost}$ and the upper limit of salary expenditure is $C_{u}^{cost}$. Moreover, since the total salary of employees can be calculated, at the next step we define the function $f_{2}$ to calculate the full salary, and the expression is:
\begin{equation}
f_{2}: T_{d a y} \sum_{j=1}^{len(K)} \sum_{i=1}^{N_{j}} \delta_{i}^{K_{j}} C_{K_{j}}
\end{equation}
\hspace*{1em} Therefore, the expression of constraint $\varphi_{k4}$  is:
\begin{equation}
\left\{\begin{array}{l}
\varphi_{k 4}: C_{l}^{\operatorname{cost}} \leq f_{2} \leq C_{u}^{cost} \\
j \in \emph{K}
\end{array}\right.
\end{equation}
\hspace*{1em} As for the constraint $\varphi_{k5}$ , since it expresses the limit of the total number of employees in each post, we need to sum up the number of employees in each post. Finally, the constraint can be expressed as:
\begin{equation}
\varphi_{k 5}: N=\sum_{j=1}^{len(K)} N_{j} \leq M
\end{equation}
\hspace*{1em}When considering the constraint $\varphi_{k6}$  , which involves employee vacation time, A counter $Count_{(\varphi=\text {True})}$  for analysis condition $\varphi$  needs to be created, and the initial value of the counter is assigned to 0. The counting process of the counter can be interpreted as increasing the value of $Count_{(\varphi=\text {True})}$ by one when $\varphi=True$ , so this counter can be used to calculate the vacation time. We also define the upper limit of employees' vacation time as  $T_{rest}$ so that $\varphi_{k6}$  can be expressed as:
\begin{equation}
\varphi_{k 6}: 0 \leq \operatorname{Count}_{\delta_{i}^{K_{j}}=0} \leq T_{\text {rest}}
\end{equation}
\indent ii.\textbf{Rules for working scheduling}\\
\indent For emergency task constraint $\varphi_{y1}$ , the general way to deal with it should be to transfer $\alpha$ employees in work and spend certain amount of time $T_{0}$  to complete the task. The priority of emergency task should be higher than that of normal work. We should also consider the mechanism of reward and punishment, these changes can be reflected in the expenditure of wages and determined by $C_{cost}$, which can be determined by the expectations of the probability distribution $E[\varphi_{y1}=True]$. Moreover, the total working time $T_{total}$ and the wage expenditure \emph{C} will change. Therefore, the constraint condition can be expressed as:
\begin{equation}
\varphi_{y 1}:\left\{\begin{array}{l}
N^{\prime}=N-\alpha \\
T^{\prime}=T_{\text {total}}-T_{0} \\
C^{\prime}=C_{\text {cost}}-C_{\text {bonus}}+C_{\text {punishment}}
\end{array}\right.
\end{equation}
\hspace*{1em}For constraint $\varphi_{y2}$ , the minimum number  $N_{l}^{K_{j}}$ and the maximum number $N_{u}^{K_{j}}$  on each job \emph{j} need to be considered. The value of  $N_{u}^{K_{j}}$ can be calculated using:
\begin{equation}
N_{u}^{K_{j}}=\frac{N_{l}^{K_{j}}}{\sum_{i=1}^{K} N_{l}^{K_{i}}} M
\end{equation}
\hspace*{1em}Therefore, $\varphi_{y2}$ can be expressed as:
\begin{equation}
\varphi_{y 2}: \forall j \in \emph{K}, \quad N_{l}^{K_{j}} \leq N_{j} \leq N_{u}^{K_{j}}
\end{equation}
\hspace*{1em}Regarding the expression of constraint  $\varphi_{y 3}$, it involves the generation algorithm of the shift table. Because of its special diversity (for example, the diversity of order and the diversity of task selection), no universal representation method has been discovered yet. When solving the problem, we can only give a solution for its special case.\\
\indent iii. \textbf{Other important rules}\\
\hspace*{1em}Firstly we study the constraint  $\varphi_{o1}$ that allows jobs shifts (multi-shifts), including job-sharing, shift work and flextime. In this case, since the four shifts for involved working time can all be arranged with different employees, the working status of the employees needs to be re-determined. Now we define that  $\delta_{i}^{j}$ represents the working status of the employee with the number \emph{i} under the \emph{$j_{th}$} working hour for every day:
\begin{equation}
\left\{\begin{array}{l}
\delta_{i}^{j}=1 \\
\delta_{i}^{j}=0\\
s.t. \sum_{j=1}^{4}\delta_{i}^{j}>0\\
\end{array}
and
 \quad\left\{\begin{array}{l}
j=1(\text { working at MOR}) \\
j=2(\text { working at  AFT}) \\
j=3(\text { working at ENV}) \\
j=4(\text { working at MID})
\end{array}\right.\right.
\end{equation}
\hspace*{1em}Under this condition, if we consider two vectors $\vec{\beta_i}=(\delta_{i}^{1},...,\delta_{i}^{4})$ and $\vec{\gamma_m}=(M O R_{m},...,MID_{m})$, the variable $T_{total}$ can be expressed as:
\begin{equation}
T_{\text {total }}=\sum_{m=1}^{K} \sum_{i=1}^{N_{m}} \vec{\beta_i} \cdot \vec{\gamma_m}
\end{equation}
\hspace*{1em}Similarly, the function $f_{3}$ for calculating the total salary will also change. After updating, the function  $f_{3}$ can be expressed as:
\begin{equation}
f_{3}: \sum_{n=1}^{T_{d a y}} \sum_{j=1}^{len(K)} \sum_{i=1}^{N_{j}}\left(\sum_{m=1}^{4} \delta_{m}^{j} C_{m}\right)
\end{equation}
\hspace*{1em} In this expression, $C_{m}$ corresponds to the $m_{th}$ hour’s wage. The requirement constraint $\varphi_{o2}$ for a cooperative task can either be expressed as ($\varphi_{y1}$), or we can design a solution for it separately.\\
\subsection{Scheduling Table Generation Algorithm}
Considering the automaticity requirement of generating scheduling list, this paper presents a scheduling list generator as an algorithm (See \textbf{Algorithm1}) to produce the scheduling table. The generator based on random selection of employees can guarantee the relative unbiasedness of average working time. 
\begin{breakablealgorithm}
\caption{Scheduling Table Generation Algorithm} 
\begin{algorithmic}
 \STATE $MemberList \leftarrow [person_{1},person_{2},...,person_{n}]$
 \STATE $Workable   \leftarrow [$ $] $
 \STATE $Day \leftarrow number$
 \STATE $Worktime \leftarrow [$ $]$
 \STATE $i \leftarrow 0$
\end{algorithmic}
\begin{algorithmic}
 \WHILE{$i < 7$}
 \STATE $count \leftarrow 0$
 \STATE $list \leftarrow [$ $]$
 \WHILE{$count <= n$}
 \STATE{$man \leftarrow random(MemberList) $}\\
 \IF{$man$ $\textbf{is}$ $Sutiable(Workable)$}
 \STATE $list.append(man)$
 \STATE $workable[man] \leftarrow workable[man]+ 1$
 \STATE $count \leftarrow count + 1$
 \ENDIF
 \ENDWHILE
 \STATE $Worktime.append(list)$
 \STATE $i \leftarrow i+1$
 \ENDWHILE
\end{algorithmic}
\end{breakablealgorithm}
The constraints on the logic are set by the selection of the function $Suitable()$.
\subsection{The Representation of Solution Space}
Consider that there are three key identifiers that distinguish different jobs and tasks in scheduling problem: employee identifiers, job identifiers, and time identifiers. Therefore, a solution space can be created, where the X-axis (channel 1) is the employee axis, representing different employees; the Y-axis (channel 2) is the time axis, which represents the circulation of four shifts in the morning, middle, evening and midnight; the Z-axis (channel 3) is the type of job axis, representing different types of work. For efficiency of solution process, we separate channels corresponding to Z-axis and then input different matrices to the algorithm part and in the next step we combine the different result matrices into the space. Therefore, the process to generate solution space can be explained as:
\begin{figure}[H] 
    	\centering
    	\includegraphics[width=6cm,height=7cm]  {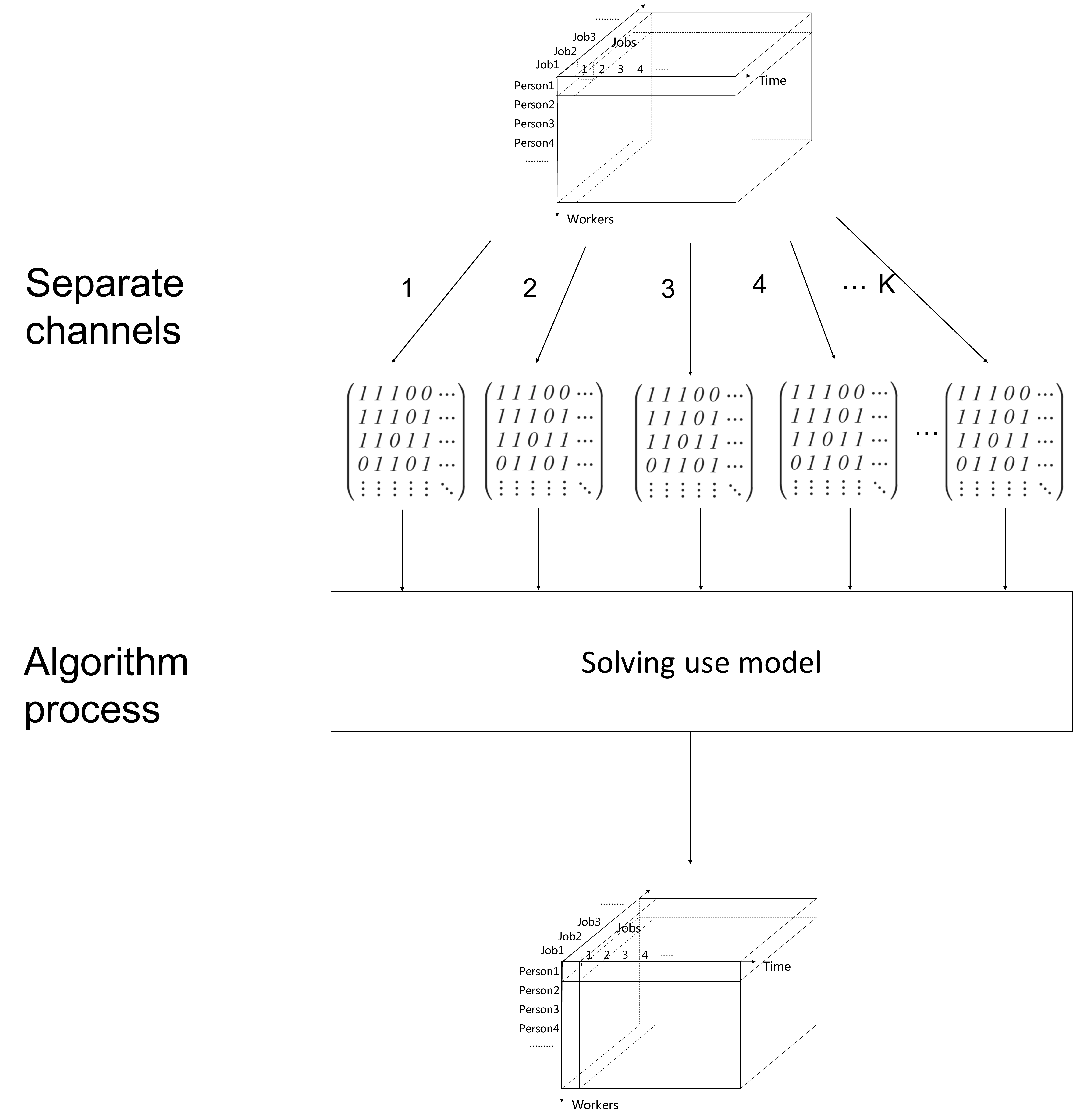}
    	\caption{Schematic Diagram of Solving Process} 
\end{figure}
\hspace*{1em}It can be proved that the separated channel matrix is equivalent to the solution space in reality. Moreover, for the elements in channel matrix rows and columns, their definition is shown as following:\\
\newline
\hspace*{1em}\hspace*{1em}$a_{(i, j)}:\left\{\begin{array}{l}i: \text { rows mean the number of staff } \\ j: \text { columns mean working time } \\ a_{(i, j)}=1 \quad \text { this worker attends } \\ a_{(i, j)}=0 \quad \text { this worker does not attend }\end{array}\right.$
\subsection{Evolution\&Genetic Algorithm Implementation}
In this paper, inspired by the work finished by Wang, Yalaoui, and Dugardin \cite{wang2017genetic}, two improved genetic algorithms based on the adaptive multi-dimensional input are used to solve the scheduling problem.\\
\hspace*{1em}For the single-objective optimization problem, we use an improved differential evolution algorithm to solve the problem. The idea of the evolution algorithm is derived from natural selection and evolutionary in biology, and it is based on genetic algorithm. Its convergence and robustness are also well suited for the solution of MSPs. The flowchart of our designed evolution algorithm (EA) is shown below:
\begin{figure}[H] 
    	\centering
    	\includegraphics[height=7cm,width=7cm]{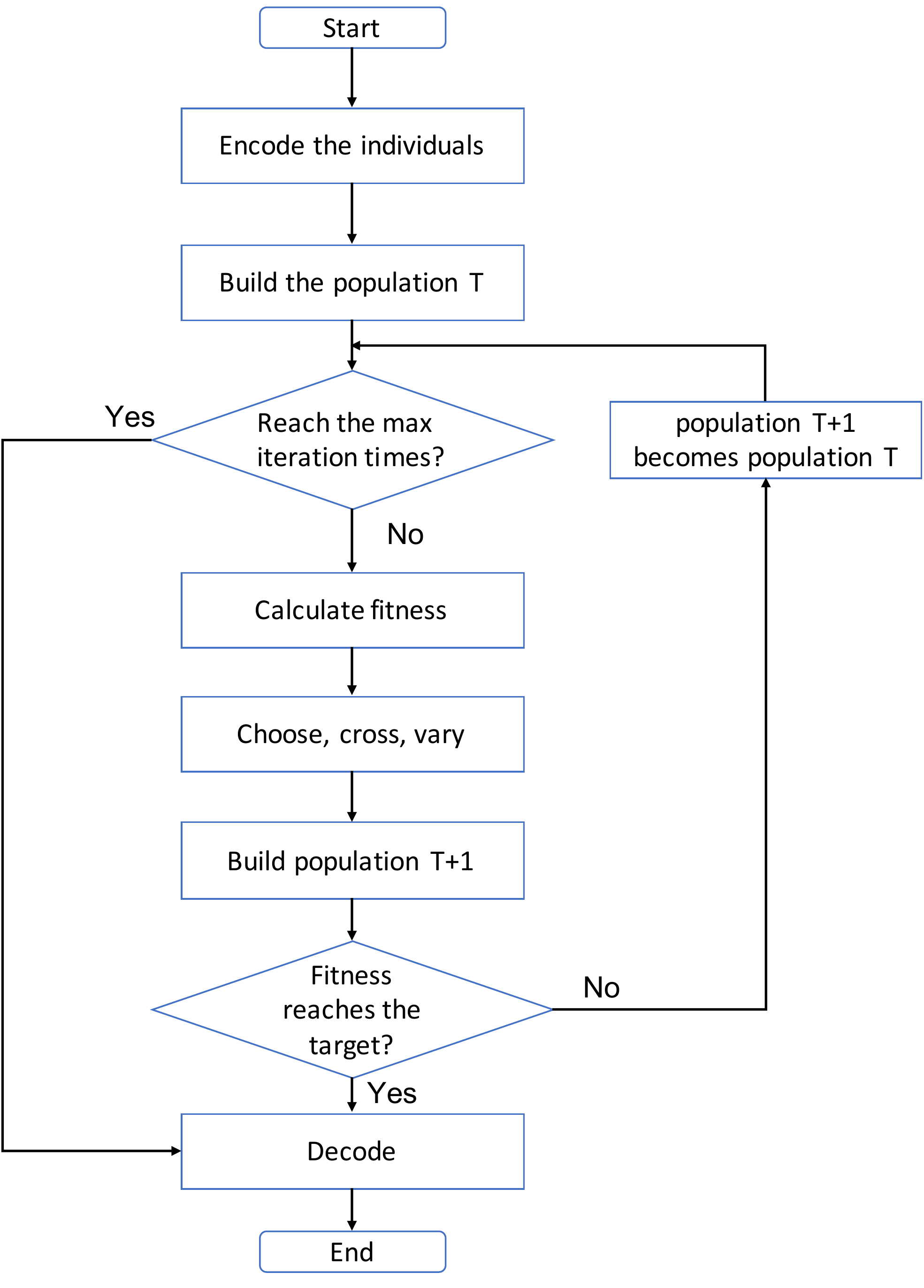}
    	\caption{Improved EA Flowchart} 
\end{figure}
\hspace*{1em}The algorithm consists of seven parts: initialization of the population, calculation of adaptation, selection, cross, vary, evolution and decode. Here are the details for our description:\\
\begin{itemize}
\item \textbf{Initialization of the population}: Initialization of the population means coding and entering the independent variables to be optimized into the population matrix. In this step, we have adopted an optimal coding scheme, and refer to each member of the population as an individual.
\item \textbf{Calculation of adaptation}: calculation of adaptation refers to a measure of the degree of adaptation for different individuals. The degree of adaptation of an individual is the difference between the value of the function to which the individual belongs and the optimal solution. The greater the degree of adaptation, the closer the individual is to the optimal solution.
\item \textbf{Selection}: Selection refers to the selection of some individuals in a population for elimination based on certain probability. Where the probability of selecting individuals is determined by the magnitude of adaptation. In this design, we will use advanced selecting method, which is based on the choices for different low level selection method such as rows, tour, etc.
\item \textbf{Cross and vary}: Cross and vary are key steps in achieving optimization of genetic algorithms. Cross refers to the fact that components can be exchanged between different individuals after encoding. Vary refers to the fact that the value of a single individual can be mutated during the optimization process.
\item \textbf{Evolution}: Evolution refers to the adaptation analysis step, which is the process of comparing with fitness in the flowchart. The comparison allows us to find the individuals we need, and if we don't find the optimal solution after one comparison, the evolution algorithm will generate a new population and continue iterating until our needs are met.
\item \textbf{Decode}: Decode refers to restoring the encoded optimal solution to its original value.
\end{itemize}
\hspace*{1em} This algorithm is convergent and we will generate the stable solution by iterating. In addition, we will provide a more detailed introduction in the experiment section of this paper.\\
\hspace*{1em} For the multi-objective optimization problem, the required combinatorial optimization can still be solved by using the model proposed in this paper. We use a modified GA with elite strategy inspired by other researches \cite{deb2002fast}.\\
\hspace*{1em} The GA with an elite strategy draws on the idea of Pareto optimization and defines two different types of individuals: dominant individuals and nondominant individuals. A dominant individual is a relationship between two individuals $i$ and $j$. If $i$ has larger fitness than $j$, then we can say that individual $i$ dominates individual $j$. A nondominant individual is "an individual that is not dominated by any member of the population", which means the individual that is most adaptive (same as the optimal solution we are looking for).\\
\hspace*{1em} According to the principle of Pareto optimization, we need to sort the adaptations of the individuals in the population to find non-dominant individuals after each iteration that produces a new population. We also need to define $crowd(u,v)$ by the value of the function corresponding to the individual, i.e. the difference between the function value corresponding to the $u_{th}$ individual and the ${v}_{th}$ individual. During the iterative process, we need to gradually eliminate dominant individuals that are more crowded compared to non-dominant individuals. When the fitness and number of nondominant individuals reach our expect, we can stop the interaction steps. If the function optimization result is convergent, then the number of nondominant individuals is also maximized, which means that the multi-objective optimization function has also obtained a Pareto optimal solution. The flowchart of the GA with elite strategy is as follows:
\begin{figure}[H] 
    	\centering
    	\includegraphics[height=6.5cm,width=7cm]{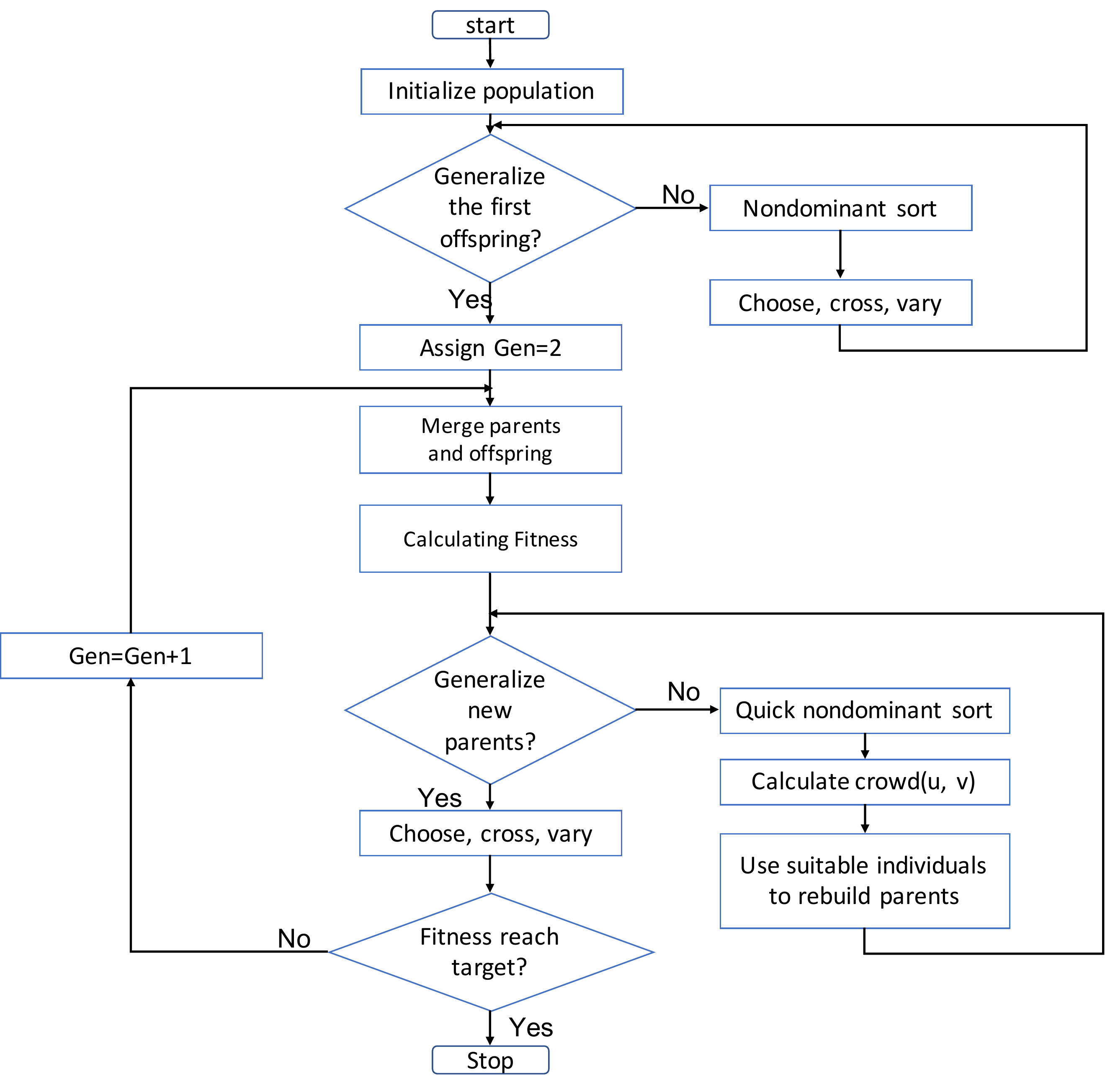}
    	\caption{GA with Elite Strategy} 
\end{figure}
\hspace*{1em} This paper provides some innovations in the coding as well as evolutionary aspects. We allow the evolution algorithm to accept RI mixed code or binary code inputs and compare the differences between the two methods. On the evolutionary side we use a penalty function approach to make the optimal solution satisfy the constraints requirement.\\
\hspace*{1em}For the representation of constraints, it is necessary to introduce the mechanism of penalty function to combine the constraint conditions with EA in the process of solving, and there are mainly two approaches to express the penalty function: external penalty function method \cite{shahnazari2013solving} and internal penalty function method \cite{du2007optimization}. Both methods are tried in this paper and it is proved that both methods can be used to solve manpower scheduling problem.\\
\hspace*{1em}To describe the features of different penalty functions more precisely, we make two graphs to simulate the working process for different penalty functions.\\
\hspace*{1em}Internal penalty functions can only be applied to solve problems with inequality constraints, since the basic rules of internal penalty functions are to set high penalty at position near the boundary for the optimization process.
\begin{figure}[H] 
    	\centering
    	\includegraphics[height=5cm,width=7cm]{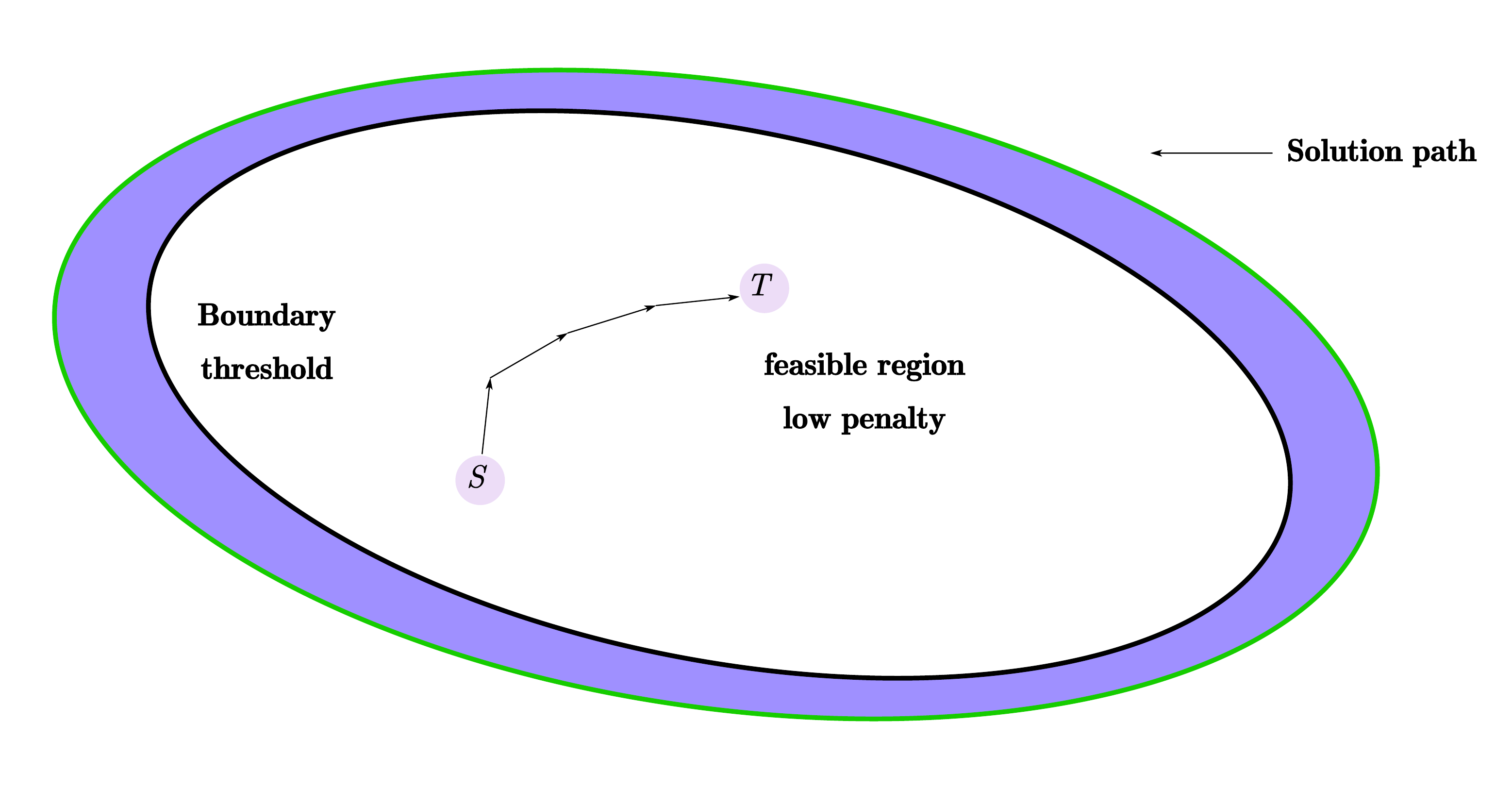}
    	\caption{Internal Penalty Function} 
\end{figure}
External penalty functions can penalize those target points that attempt to violate the constraints in the process of solving unconstrained problems so that it will finally solve an unconstrained problem.
\begin{figure}[H] 
    	\centering
    	\includegraphics[height=6cm,width=7cm]{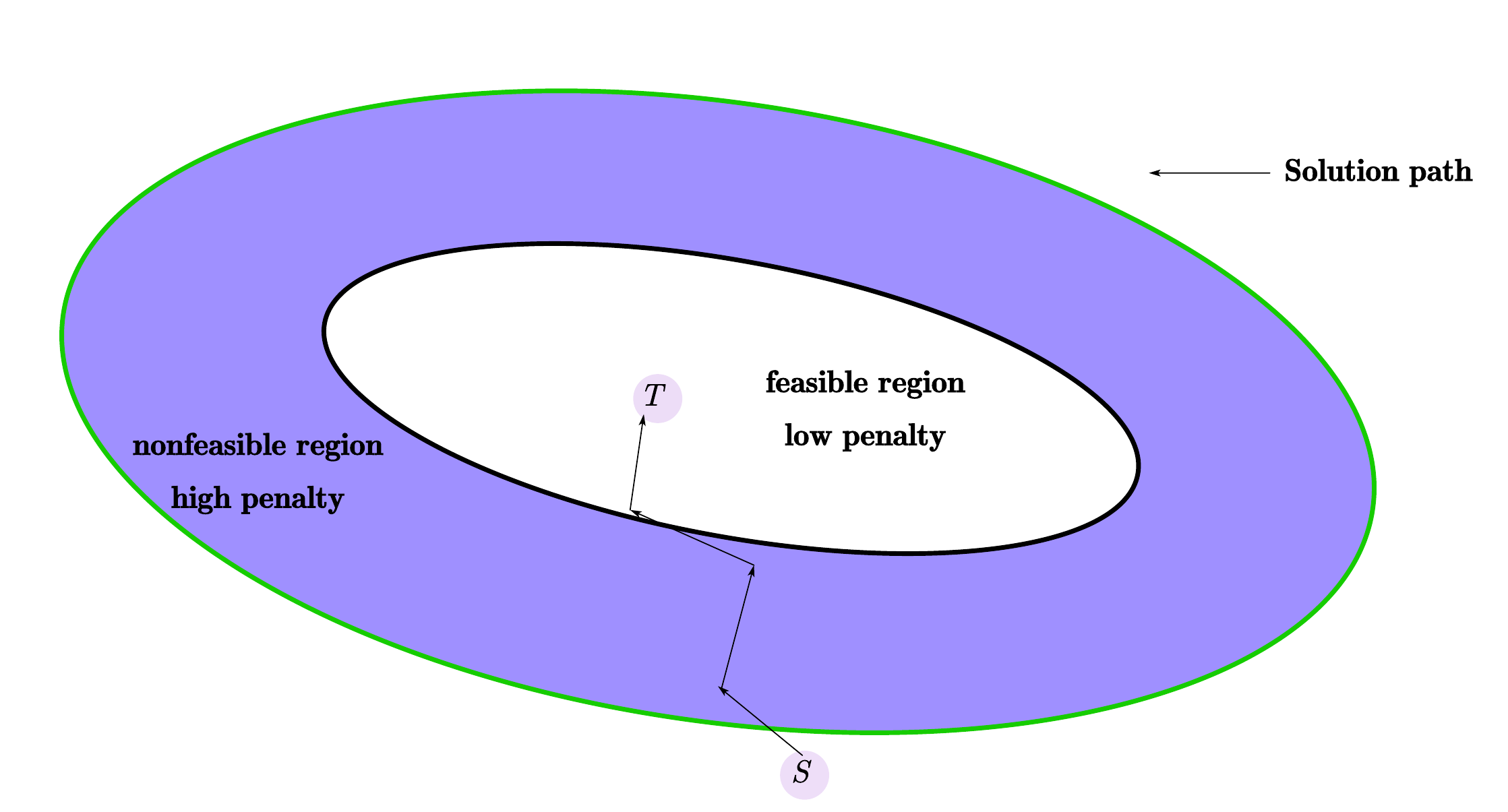}
    	\caption{External Penalty Function} 
\end{figure}
\section{Experiments and Tests of the Model}
The following experiment part is about how to apply the new scheduling model proposed in this paper to fit one requirement in a shopping mall. $M$ is assigned to 60. These experiments are run on a PC with Intel core, i5-5257u, 2.7-3.1 GHz.
\begin{table}[H]
\caption{Employees List}
\begin{center}
\setlength{\tabcolsep}{1mm}{
\begin{tabular}{|c|c|c|c|c|c|c|}
	\hline
	Code&a&b&c&d&e&f\\ \hline
	Type&manager&clerk&guard&salesclerk&tallyclerk&cleaner\\ \hline
\end{tabular}}
\end{center}
\end{table}
\subsection{Experiment1: Minimize Total Time under the Basic Constraints}
To minimize total time under the constraints $\varphi_{k 1} \wedge \varphi_{k 2} \wedge \varphi_{k 3} \wedge \varphi_{k 4} \wedge \varphi_{k 5} \wedge \varphi_{k 6}$ , using the preprocess analysis for this problem, the requirement can be expressed as:
\begin{equation}
\left\{\begin{array}{ll}
\min & T_{\text {total}} \\
\text {s.t.} & \varphi_{k 1} \wedge \varphi_{k 2} \wedge \varphi_{k 3} \wedge \varphi_{k 4} \wedge \varphi_{k 5} \wedge \varphi_{k 6}
\end{array}\right.
\end{equation}
\hspace*{1em}The initial number of individuals in the population is 100. To compare the efficiency of EA, in this experiment, two different chromosome coding methods are adopted to solve the minimum value. They are binary coding and real \& integer mixed coding.\\
\hspace*{1em}If BG (binary coding) method is adopted, and after 50 generations, min value of this objective function is 5250. For the optimal solution, the values of the independent variables are:
\begin{figure}[H]
\begin{minipage}{0.55\linewidth}
 \centerline{\includegraphics[width=5cm,height=4cm]{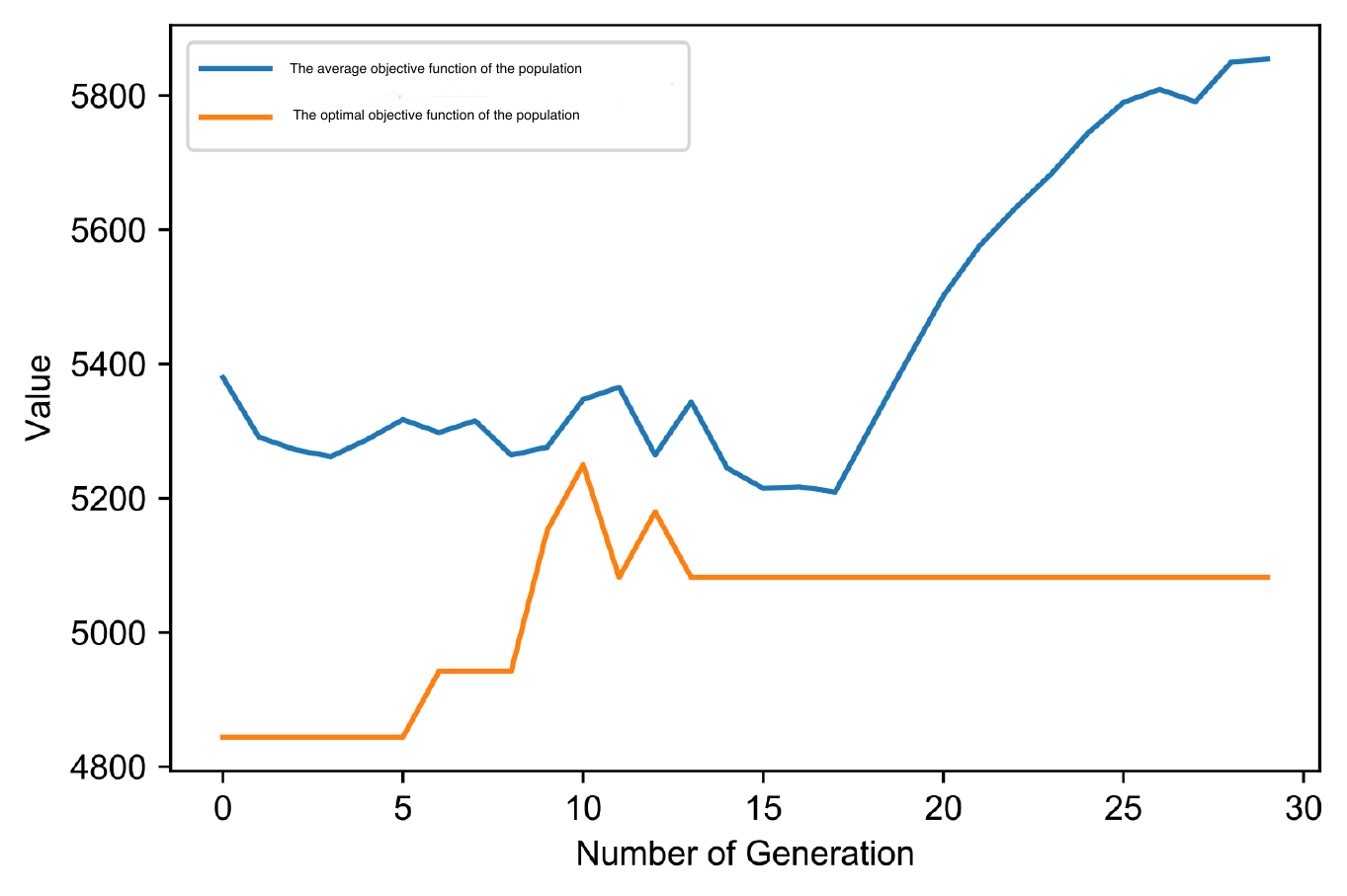}}
\end{minipage}
\hfill
\begin{minipage}{0.4\linewidth}
\begin{equation}\nonumber
\left\{\begin{array}{l}
a=3 \\
b=10 \\
c=4 \\
d=8 \\
e=7 \\
f=8
\end{array}\right.
\end{equation}
\end{minipage}
\caption{BG Coding Result}
\end{figure}
If RI (real number \& integer number mixed coding) is adopted, and after 50 generations, the EA will get a minimum value of 4942. For the optimal solution, the values of the independent variables are:
\begin{figure}[H]
\begin{minipage}{0.55\linewidth}
 \centerline{\includegraphics[width=5cm,height=4cm]{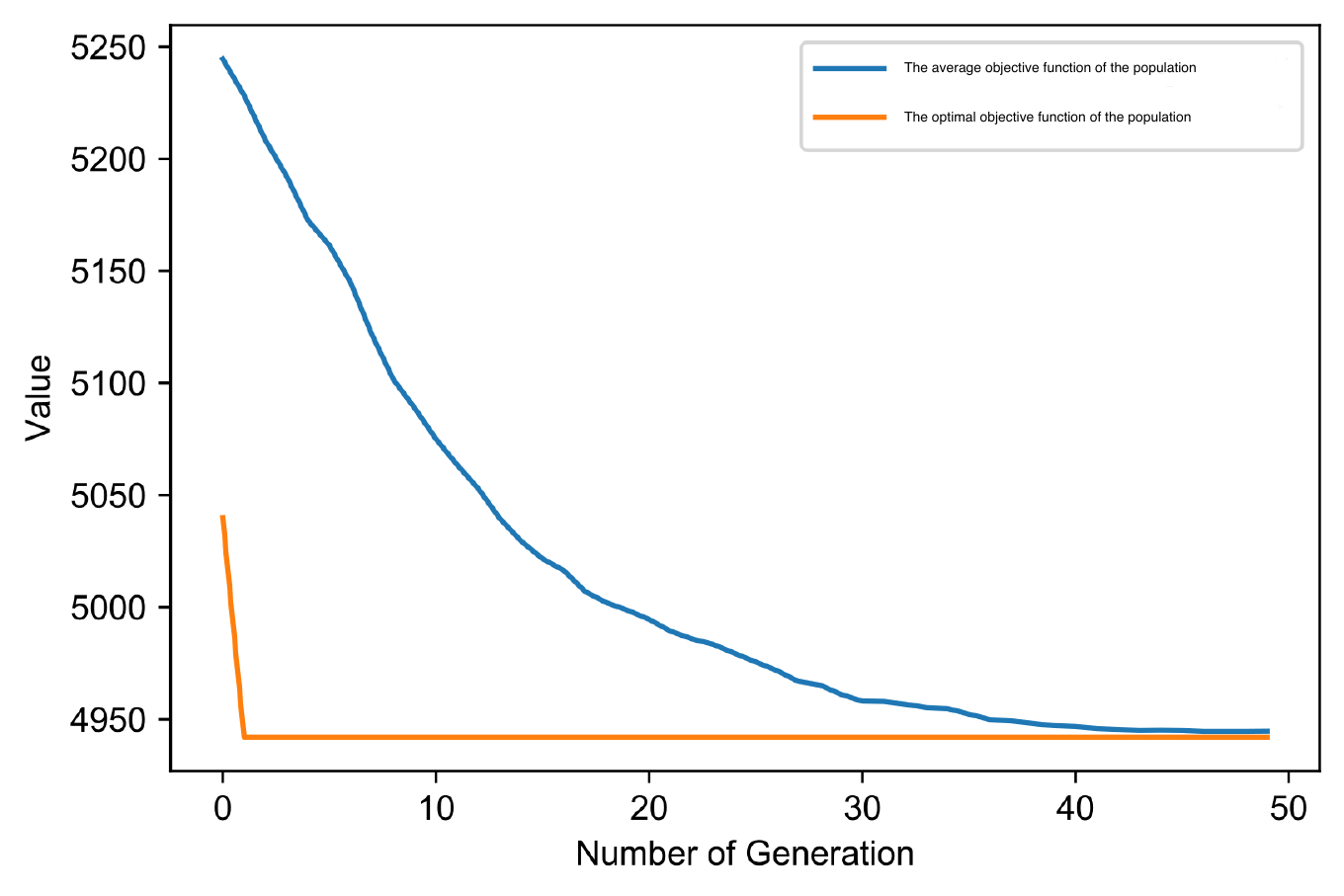}}
\end{minipage}
\hfill
\begin{minipage}{0.4\linewidth}
\begin{equation}\nonumber
\left\{\begin{array}{l}
a=3 \\
b=10 \\
c=3 \\
d=8 \\
e=6 \\
f=8
\end{array}\right.
\end{equation}
\end{minipage}
\caption{RI Coding Result}
\end{figure}
From the experimental results, we can see that the result obtained by using RI coding has a faster convergence speed, and the objective function value is relatively smaller, which shows that RI coding is more suitable for the following solving operations.\\
\hspace*{1em}After obtaining the number of people needed for the arrangement, the value of the impact function should be determined. At this point, the problem has been transformed into a 0-1integer programming problem, which can be solved by using EA again, and the following results can be obtained:
\begin{table}[H]
	\small
	\caption{Experiment results obtained by EA}
	\centering
	\begin{tabular}{|c|c|c|c|c|c|c|}
		\hline
		 Evaluating times&Time cost&Optimal function value\\ \hline
		\textbf{8400}&\textbf{0.07379s}&\textbf{3850}\\ \hline
	\end{tabular}
\end{table}
We input the results obtained from above EA to the generator algorithm for scheduling table to get solution space $S_{xyz}$:
\begin{figure}[H] 
    	\centering
    	\includegraphics[height=4cm,width=5.5cm]{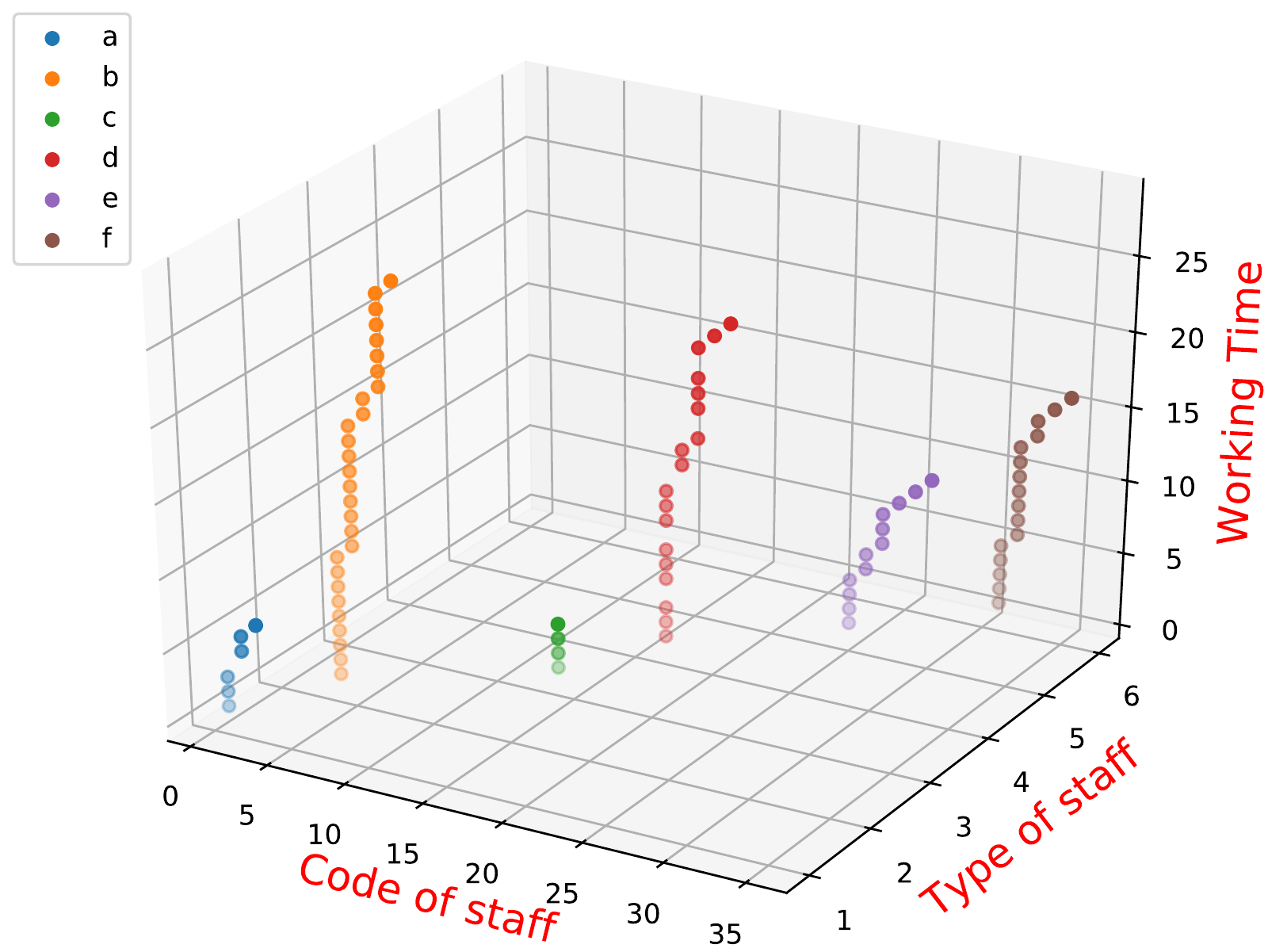}
    	\caption{Solution Space for Experiment1} 
\end{figure}
\subsection{Experiment2: Work Shifts in the Scheduling Problem}
If the work shifts ($\varphi_{o1}$) is allowed, that is, the form of which is generally determined by the number of persons required and the number of shifts required. For example, three positions in five shifts means that five people should be arranged in three positions in a shift cycle. Since the situation has changed, an algorithm can be designed to consider the situation of shifts work. The sequence of employees in the list can be adjusted by the function $Order()$. Therefore, the result of the generator algorithm is as follows:
\begin{figure}[H] 
    	\centering
    	\includegraphics[height=4cm,width=5.5cm]{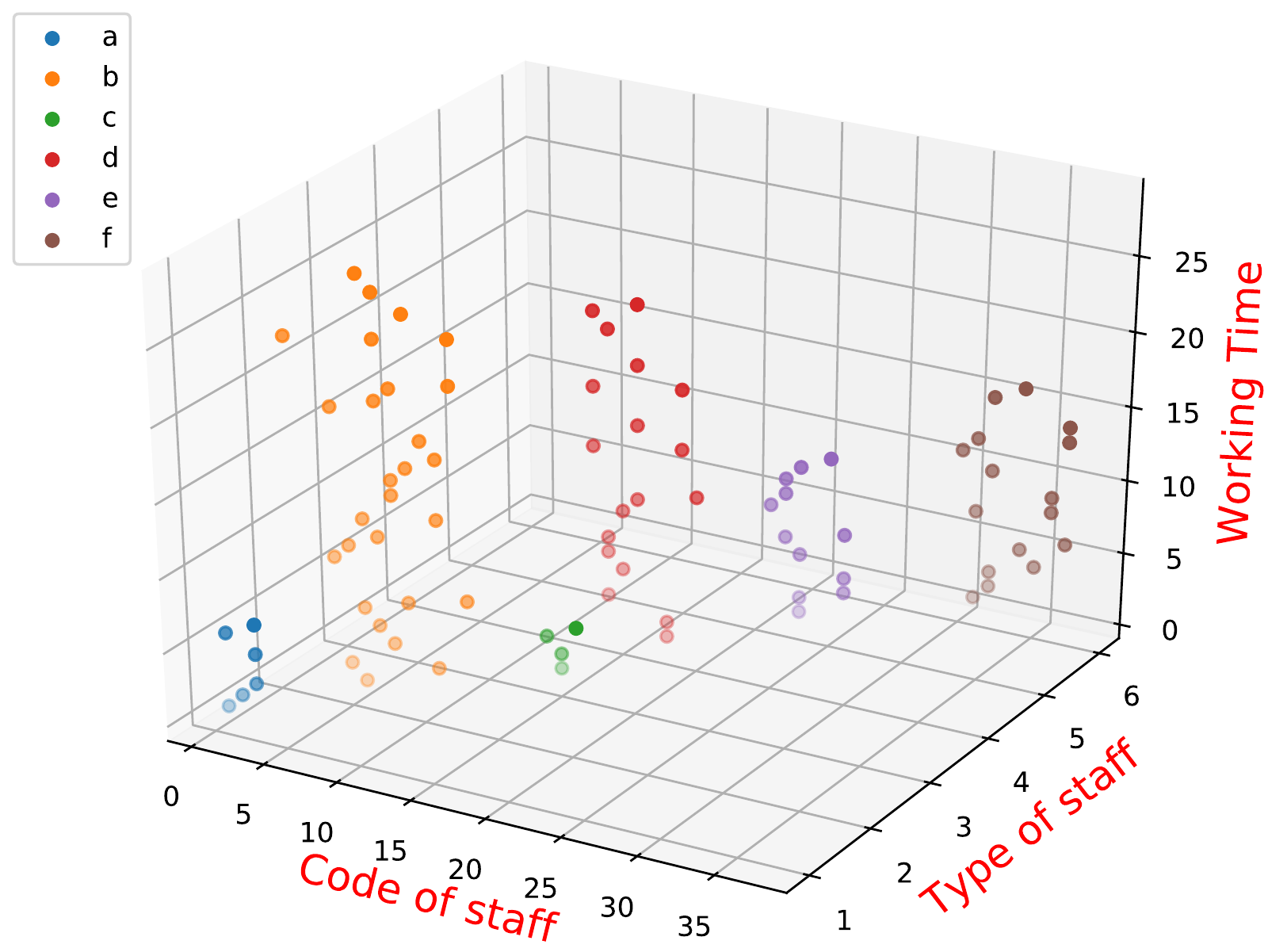}
    	\caption{Solution Space $S_{xyz}$ for Experiment2} 
\end{figure}
\subsection{Expreiment3: The Solution for the Problem after Complex Combination of Constraints.}
The experiment is divided into two parts to explore the possibility of logical paradigms considering the representation of constraints to be joined by logical operators. In the first part, the penalty function is directly linked by logical operators from the point of view of evolution algorithm. For example, the constraint conditions are: each post cannot be interleaved with duty, and each post has someone on duty every day, and the total salary of the employees should be greater than a certain amount; or the employee's working time is within a certain range and the problem of the minimum total working time. The problem can be shown as:
\begin{equation}
\left\{\begin{array}{ll}
\min & T_{\text {total}} \\
\text {s.t.} & \varphi_{k 1} \wedge \varphi_{k 2} \wedge\left(\neg \varphi_{k 5}\right) \vee \varphi_{k 3}
\end{array}\right.
\end{equation}
\hspace*{1em}This graph represents the results obtained using EA:
\begin{figure}[H] 
    	\centering
    	\includegraphics[height=4cm,width=5.5cm]{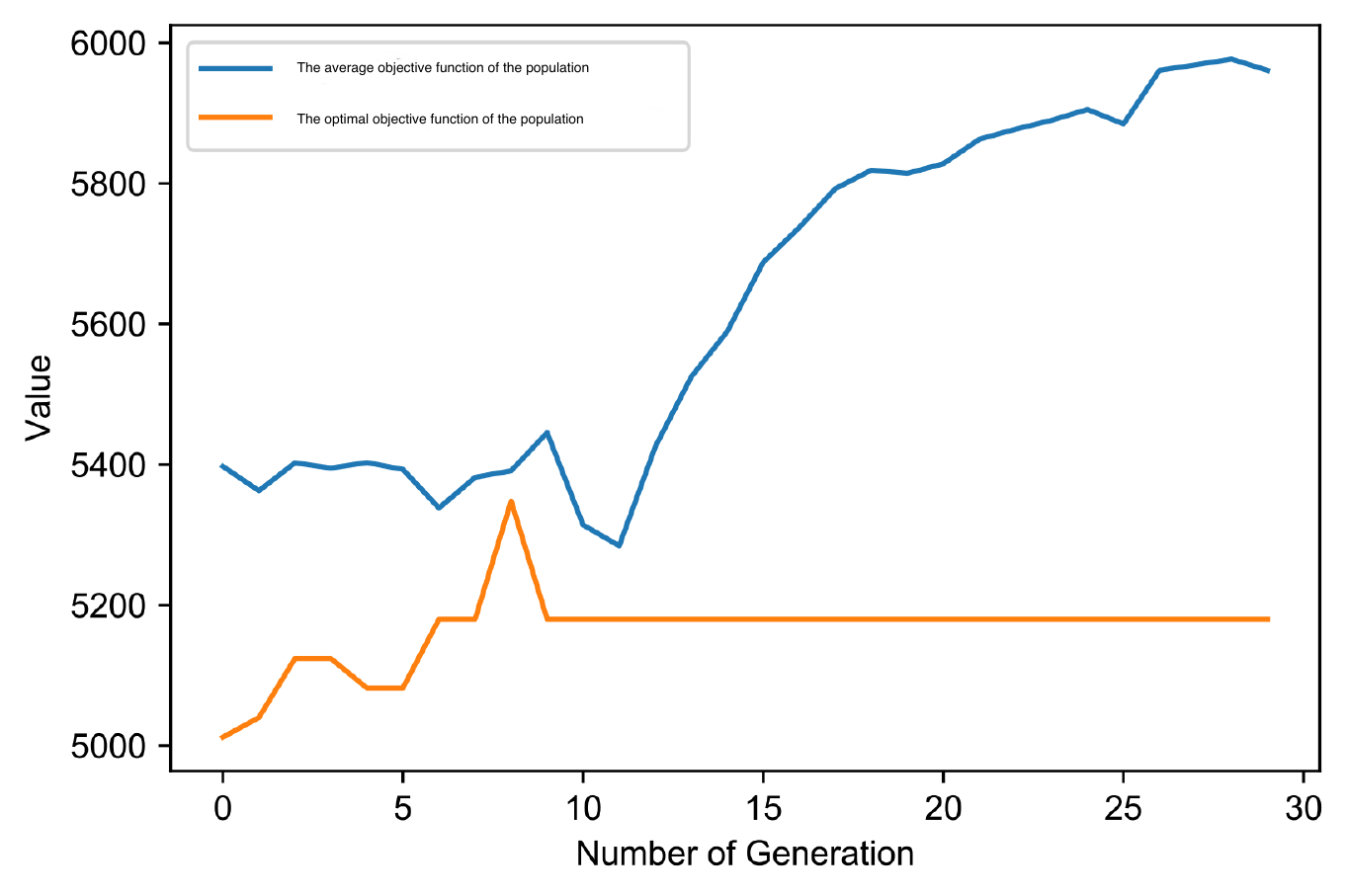}
    	\caption{Result for Experiment3} 
\end{figure}
\subsection{Experiment4: Urgent tasks contained in the scheduling problem}
Consider changes used for scheduling in the event of urgent tasks (such as collapsing shelves, theft from stores, etc.). Especially, for the collapse of shelves, employees involved are tallymen and salespersons, then the problem is transformed into:
\begin{equation}
\left\{\begin{array}{ll}
\min & T_{\text {total}} \\
\text {s.t.} & \varphi_{k 1} \wedge \varphi_{k 2} \wedge \varphi_{k 3} \wedge \varphi_{k 4} \wedge \varphi_{k 5} \wedge \varphi_{k 6} \wedge \varphi_{y 1}
\end{array}\right.
\end{equation}
\hspace*{1em}After modifying the penalty function and calling the evolution algorithm for single-objective optimization, we can get the following results:
\begin{figure}[H]
\begin{minipage}{0.48\linewidth}
 \centerline{\includegraphics[width=5.0cm,height=3.5cm]{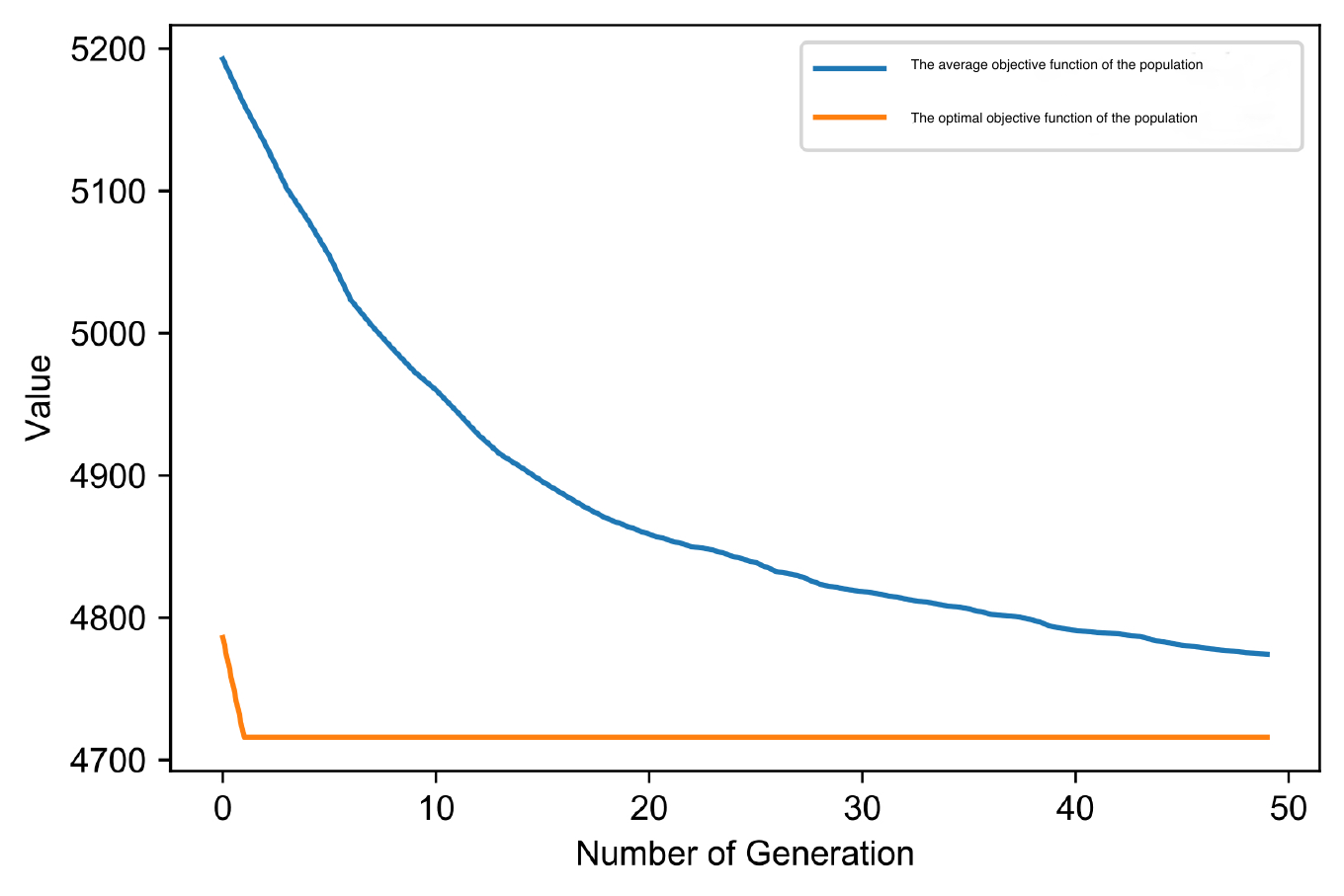}}
\end{minipage}
\hfill
\begin{minipage}{.48\linewidth}
\flushleft
\begin{equation}\nonumber
\left\{\begin{array}{l}
a=4 \\
b=11 \\
c=3 \\
d=8 \\
e=6 \\
f=8
\end{array}\right.
\end{equation}
\end{minipage}
\caption{Result for Experiment4}
\end{figure}
\subsection{Experiment5: Scheduling Problem under Multi-objective Optimization}
In reality, different enterprises and different managers will have different demands, which also determine that there are various combinations of distinct objective functions. This experiment shows the process of solving multi-objective optimization using our model. \\
\hspace*{1em}The following example shows the application of the algorithm in solving the multi-objective optimization scheduling problem. The objective functions are: Minimizing the number of clerks, guards and tallyclerks $f_{a}$ while minimizing the working time $f_{b}$. 
\begin{equation}
\left\{\begin{array}{ll}
\min & f_{a} : \sum_{i=2,3,5}{N_{i}}\\
\min & f_{b} :  \sum_{j=1}^{len(K)}{\sum_{i=1}^{N_{j}} T_{K} \delta_{i}^{K_{j}}}\\
\text {s.t.} & \varphi_{k 1} \wedge \varphi_{k 2} \wedge \varphi_{k 3} \wedge \varphi_{k 4} \wedge \varphi_{k 5} \wedge \varphi_{k 6}
\end{array}\right.
\end{equation}
After the optimization conditions are substituted in, the following results can be obtained:
\setlength{\tabcolsep}{1mm}{
\begin{table}[H]
	\caption{The parameters obtained by GA}
	\centering
	\begin{tabular}{|c|c|c|}
		\hline
		 Time cost&Optimal value of $f_{a}$&Optimal value of $f_{b}$\\ \hline
		\textbf{0.1096s}&\textbf{21}&\textbf{4844}\\ \hline
	\end{tabular}
\end{table}
}
\section{Discussion}
In order to verify the efficiency of the new model we proposed, this paper compares some baseline models with the approach proposed in this paper from two aspects. Moreover, the superiority of the new approach proposed in this paper is judged by studying the time cost, the accuracy of solutions, and the rationality of the generated results.\\
\subsection{Obtaining the Number of Employees}
In order to compare the efficiency of the solution algorithm horizontally, this paper also uses integer programming algorithm (\textbf{IP}), particle swarm optimization (\textbf{PSO}) and simulated annealing algorithm (\textbf{SA}) to complete Experiment 1 as comparison. Here are details for the baseline models:\\
\subsubsection{IP}
Integer programming refers to planning in which the variables (all or part) are restricted to integers. In this problem, the variables $x_{i}$ are restricted to integers and the constraints can be transformed into linear restrictions. Therefore we express IP method as:
\[
\min T_{total}
\]
\begin{equation}
s.t. \begin{cases}
\phi_{k1} \wedge \phi_{k2} \wedge \phi_{k3} \wedge \phi_{k4} \wedge \phi_{k5} \wedge \phi_{k6} \\ 
x_{i}=1\  or\  0\\
\end{cases} 
\end{equation}
\hspace*{1em}Moreover, we can use spicy library to solve this optimization problems with constraints.
\subsubsection{PSO}
The particle swarm algorithm is derived from the behavioral study of bird predation, and the basic idea is to search for optimal solutions through collaboration and information sharing between individuals in the group. By designing particles with only two variables, position and velocity, we can simulate the motion of the particle swarm as a search for an optimal solution in space. As one particle reaches the position of the global optimal solution $(p_{best},q_{best})$, the rest of the particles adjust their velocities according to that particle's position and eventually obtain the convergent optimal solution.\\
\hspace*{1em}The updating of velocity and position for the particle $i$ will follow these rules:
\begin{equation}
\begin{cases}
v_{i}=\omega v_{i}+c_{1}rand()( p_{besti}-x_{i}) +c_{2}rand()(g_{besti}-x_{i})&\\
 x_{i}=x_{i}+v_{i}&
 \end{cases}
\end{equation}
\hspace*{1em}$\omega$ means inertial factor, which is set as a dynamic parameter. $c_{1}$ and $c_{2}$ are learning factors. $rand()$ is a function to generate random value. It is suitable for us to adjust the value of these parameters to solve the problems. Moreover, the flowchart for PSO algorithm can be expressed as:
\begin{figure}[H] 
    	\centering
    	\includegraphics[height=9cm,width=7cm]{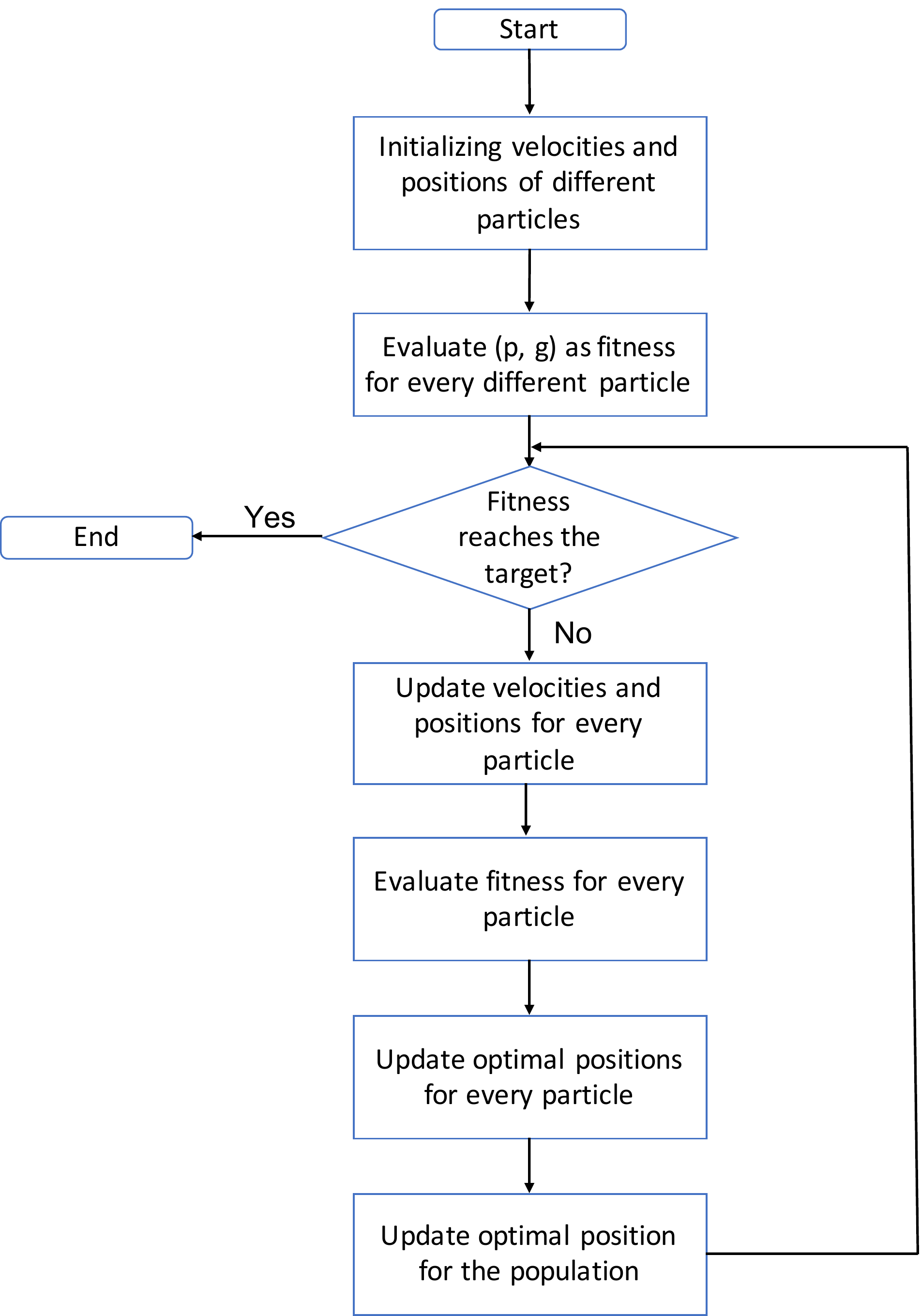}
    	\caption{Workflow for PSO} 
\end{figure}
\subsubsection{SA}
Simulated annealing algorithm is a solution based on knowledge of energy and systems to simulate the cooling process of a crystal in order to obtain an optimal solution. Annealing itself is a physics term that means "a slow decrease in temperature". From an algorithmic point of view, annealing is the process of searching for an optimal solution. For the procedure of SA, firstly we need to determine an initial temperature $T_{0}$ and secondly we need to set the working rate v of the annealing process. In addition, the unit of optimization in the simulated annealing algorithm is the state. After one iteration of state A, we get a new state B. For states A and B, the simulated annealing algorithm accepts state updates with the following probability P.
\begin{equation}
P=\left\{\begin{array}{c}1,  E(B)<E(A) \\ 
e^{-\frac{E(B)-E(A)}{T},  E(B) \geq E(A)}\end{array}\right.
\end{equation}
\hspace*{1em}$E(A)$ and $E(B)$ represent the energy of state A and B. $T$ means temperature.\\
\hspace*{1em}After multiple state transfers, if the objective function converges, we can obtain a state with the lowest global energy (i.e., the termination temperature $T_{f}$. The value of the independent variable corresponding to this temperature (energy) is also the optimal solution under the simulated annealing algorithm. The flowchart of SA process can be expressed as:
\begin{figure}[H] 
    	\centering
    	\includegraphics[height=7cm,width=6cm]{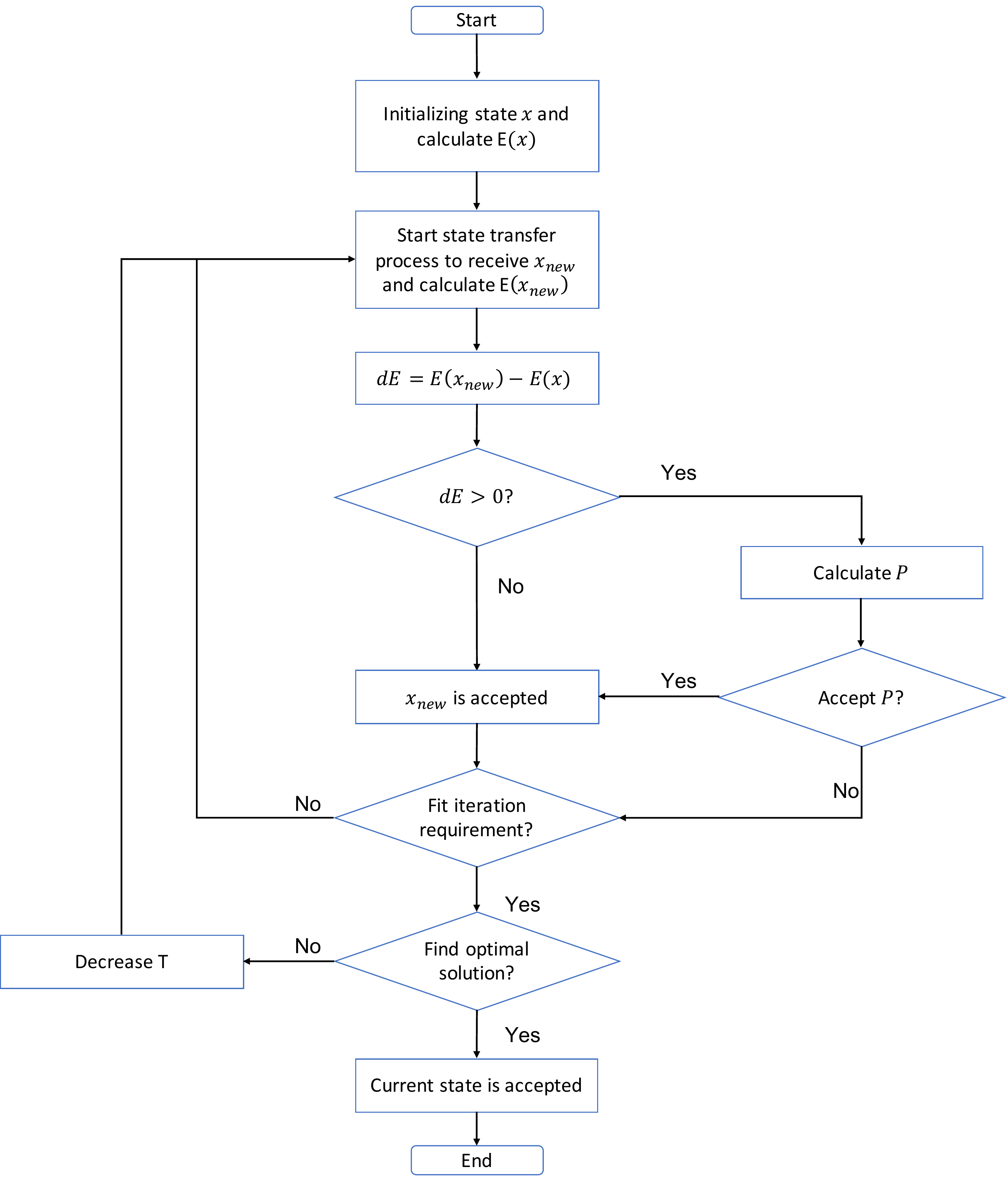}
    	\caption{Workflow for SA} 
\end{figure}
\hspace*{1em}Here are the results for testing experiment1 using different algorithm:
\begin{table}[H]
\caption{Comparison Results}
\begin{center}
\setlength{\tabcolsep}{0.4mm}{
\begin{tabular}{|c|c|c|c|}
		\hline
		 Name of algorithm&Time cost&Accuracy&Convergence\\ \hline
		\textbf{New-EA}&\textbf{0.19899s}&\textbf{100\%}&\textbf{2}\\ \hline
		IP&0.01378s&70\%&1\\ \hline
		PSO&0.26791s&100\%&3\\ \hline
		SA&1.07792s&83\%&4\\ \hline
\end{tabular}}
\end{center}
\end{table}
\hspace*{1em}Firstly, we explain the meaning of the data in the table. $Running time$: This variable refers to the time cost by different algorithms. $Accuracy$: This value is defined as $\frac{Result\quad of\quad New-EA}{Result\quad of\quad IP/PSO/SA} * 100\%$. $Convergence$: Probability of a stable solution. The value is defined as $rank(\frac{\#stable\quad result}{\#total\quad experiment} * 100\%)$.\\ 
\hspace*{1em}Compared with other algorithms, the new evolution algorithm proposed in this paper has apparent advantages, which are the high efficiency and the outstanding performance in selecting optimal solutions. In the running time, it is 25.7\% faster than the PSO; in the accuracy of the running result, it is 17\% higher than the SA. Therefore, the evolution algorithm proposed in this paper is useful in completing complex arrangements manpower scheduling task, and it is worth further optimization and investigation.
\subsection{Generator Algorithm of Scheduling Plan}
In comparing the scheduling table generation algorithm, this paper selects a manpower scheduling algorithm related to the research results of Korf \cite{korf2002new} for comparison, which is adapted by the baseline model \cite{project1} that has related code collections on Github. It can be used to study the effectiveness of the generation algorithm proposed in our paper. By comparing the completion of the two algorithms' scheduling tasks for the two different periods of 7 days and 30 days, we can discuss the effective solutions shown in Figure 14\&15. Therefore, the comparative experiment proves the value of our model.
\begin{figure}[H] 
    	\centering
    	\includegraphics[height=4cm,width=6cm]{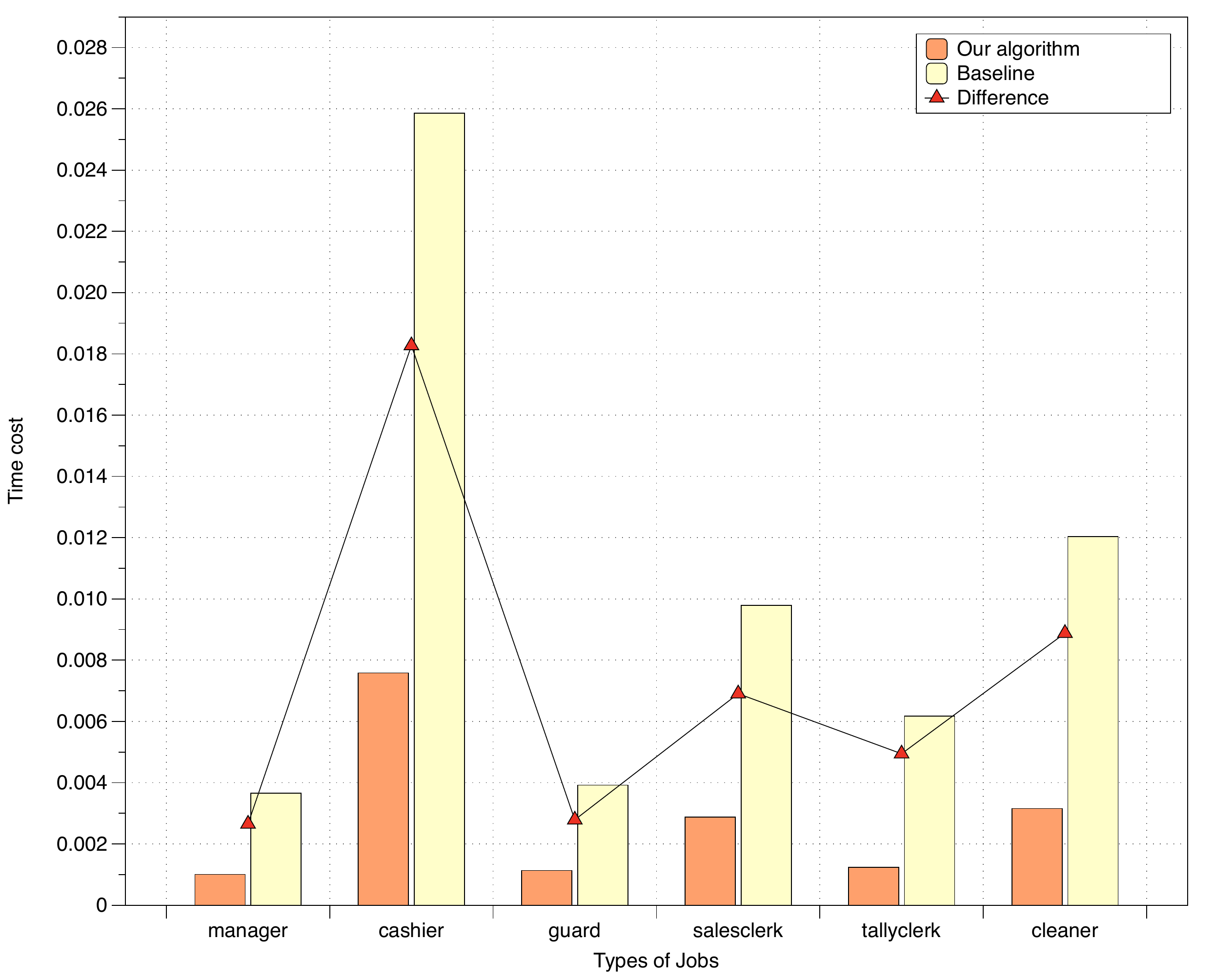}
    	\caption{Results for 7 Days} 
\end{figure}
\begin{figure}[H] 
    	\centering
    	\includegraphics[height=4cm,width=6cm]{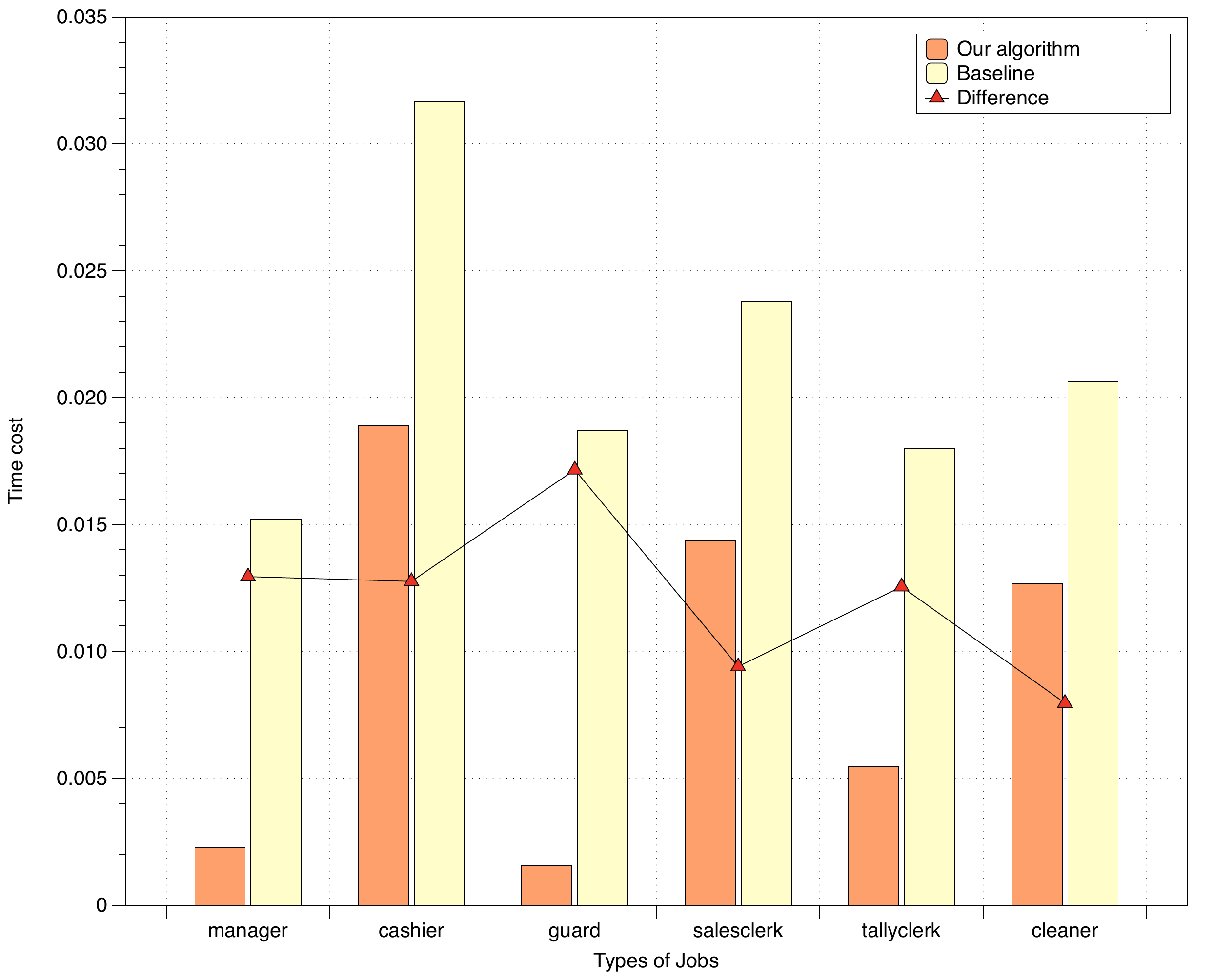}
    	\caption{Results for 30 Days} 
\end{figure}
The result is very insightful. We can determine that this paper has a significant advantage over the method provided by the baseline model in the optimization of solution efficiency from the results. For the arrangement of different jobs, the efficiency can be increased by up to 40.28\%, and at least by 28.91\%.
\section{Conclusions}
This paper proposes a new model that can be applied to solving single-shift MSP or multi-shift MSP. It is creative and improves the efficiency of enterprise scheduling tasks, thereby optimizing the enterprise's human resources structure and creating more excellent value. Our work transforms the traditional scheduling problem as a combinatorial optimization problem and creates a novel module to solve MSP. First of all, this paper combines different constraints by combining logical paradigms. Subsequently, this paper determines the number of employees required for different jobs using our improved evolution algorithm. Finally, this paper generates an arrangement table for the specific MSP. In addition, to verify the new scheduling model's efficiency, this paper compares the new model with different baseline models in terms of different aspects, such as efficiency, convergence, and accuracy. The experimental results show that the new model (new algorithm) proposed in this paper performs better than the baseline models, thus affirming our work's research value.
\section{Acknowledgments}
We are grateful to anonymous reviewers for their helpful comments. This work is supported in part by NSFC Grant No. 61772315 and 61861136012, and National Key R\&D Program of China (No. 2018AAA0101100).

\bibliography{reference}

\begin{thebibliography}{10}
\providecommand{\url}[1]{#1}
\csname url@samestyle\endcsname
\providecommand{\newblock}{\relax}
\providecommand{\bibinfo}[2]{#2}
\providecommand{\BIBentrySTDinterwordspacing}{\spaceskip=0pt\relax}
\providecommand{\BIBentryALTinterwordstretchfactor}{4}
\providecommand{\BIBentryALTinterwordspacing}{\spaceskip=\fontdimen2\font plus
\BIBentryALTinterwordstretchfactor\fontdimen3\font minus
  \fontdimen4\font\relax}
\providecommand{\BIBforeignlanguage}[2]{{%
\expandafter\ifx\csname l@#1\endcsname\relax
\typeout{** WARNING: IEEEtran.bst: No hyphenation pattern has been}%
\typeout{** loaded for the language `#1'. Using the pattern for}%
\typeout{** the default language instead.}%
\else
\language=\csname l@#1\endcsname
\fi
#2}}
\providecommand{\BIBdecl}{\relax}
\BIBdecl

\bibitem{pan2010solving}
Q.-K. Pan, P.~N. Suganthan, T.~J. Chua, and T.~Cai, ``Solving manpower
  scheduling problem in manufacturing using mixed-integer programming with a
  two-stage heuristic algorithm,'' \emph{The International Journal of Advanced
  Manufacturing Technology}, vol.~46, no. 9-12, pp. 1229--1237, 2010.

\bibitem{dantzig1954minimizing}
G.~B. Dantzig and D.~R. Fulkerson, ``Minimizing the number of tankers to meet a
  fixed schedule,'' \emph{Naval Research Logistics Quarterly}, vol.~1, no.~3,
  pp. 217--222, 1954.

\bibitem{romero2016man}
P.~Romero, F.~Robledo, P.~Bevilacqua, and M.~Delafuente, ``Man power
  optimization in large-scale corporations,'' \emph{Investigaci{\'o}n
  Operacional}, vol.~37, no.~2, pp. 173--183, 2016.

\bibitem{kumar2018load}
T.~S. Kumar, Z.~Wang, A.~Kumar, C.~M. Rogers, and C.~A. Knoblock, ``Load
  scheduling of simple temporal networks under dynamic resource pricing.'' in
  \emph{AAAI}, 2018, pp. 6227--6236.

\bibitem{sabar2012agent}
M.~Sabar, B.~Montreuil, and J.-M. Frayret, ``An agent-based algorithm for
  personnel shift-scheduling and rescheduling in flexible assembly lines,''
  \emph{Journal of Intelligent Manufacturing}, vol.~23, no.~6, pp. 2623--2634,
  2012.

\bibitem{zheng2017teaching}
H.-y. Zheng, L.~Wang, and X.-l. Zheng, ``Teaching--learning-based optimization
  algorithm for multi-skill resource constrained project scheduling problem,''
  \emph{Soft Computing}, vol.~21, no.~6, pp. 1537--1548, 2017.

\bibitem{ciancio2018integrated}
C.~Ciancio, D.~Lagan{\`a}, R.~Musmanno, and F.~Santoro, ``An integrated
  algorithm for shift scheduling problems for local public transport
  companies,'' \emph{Omega}, vol.~75, pp. 139--153, 2018.

\bibitem{ho2010solving}
S.~C. Ho and J.~M. Leung, ``Solving a manpower scheduling problem for airline
  catering using metaheuristics,'' \emph{European Journal of Operational
  Research}, vol. 202, no.~3, pp. 903--921, 2010.

\bibitem{turker2018integrated}
T.~T{\"u}rker and A.~Demiriz, ``An integrated approach for shift scheduling and
  rostering problems with break times for inbound call centers,''
  \emph{Mathematical Problems in Engineering}, vol. 2018, 2018.

\bibitem{narasimhan2000algorithm}
R.~Narasimhan, ``An algorithm for multiple shift scheduling of hierarchical
  workforce on four-day or three-day workweeks,'' \emph{INFOR: Information
  Systems and Operational Research}, vol.~38, no.~1, pp. 14--32, 2000.

\bibitem{yingjun2010external}
W.~Yingjun and S.~Mingqing, ``The external penalty function method for
  optimization design of reinforced concrete underground penstock,'' in
  \emph{2010 International Conference on Computing, Control and Industrial
  Engineering}, vol.~2.\hskip 1em plus 0.5em minus 0.4em\relax IEEE, 2010, pp.
  345--347.

\bibitem{su2017integrated}
T.-S. Su and S.-C. Liu, ``Integrated supporting cooperation model with fuzzy
  approach for staff scheduling problem in service supply chain,'' in
  \emph{2017 IEEE International Conference on Industrial Engineering and
  Engineering Management (IEEM)}.\hskip 1em plus 0.5em minus 0.4em\relax IEEE,
  2017, pp. 369--373.

\bibitem{wang2017genetic}
X.~Wang, F.~Yalaoui, and F.~Dugardin, ``Genetic algorithms hybridized with the
  self controlling dominance to solve a multi-objective resource constraint
  project scheduling problem,'' in \emph{2017 IEEE International Conference on
  Service Operations and Logistics, and Informatics (SOLI)}.\hskip 1em plus
  0.5em minus 0.4em\relax IEEE, 2017, pp. 39--44.

\bibitem{deb2002fast}
K.~Deb, A.~Pratap, S.~Agarwal, and T.~Meyarivan, ``A fast and elitist
  multiobjective genetic algorithm: Nsga-ii,'' \emph{IEEE transactions on
  evolutionary computation}, vol.~6, no.~2, pp. 182--197, 2002.

\bibitem{shahnazari2013solving}
P.~Shahnazari-Shahrezaei, R.~Tavakkoli-Moghaddam, and H.~Kazemipoor, ``Solving
  a multi-objective multi-skilled manpower scheduling model by a fuzzy goal
  programming approach,'' \emph{Applied Mathematical Modelling}, vol.~37,
  no.~7, pp. 5424--5443, 2013.

\bibitem{du2007optimization}
X.~Du, H.~Wang, Q.~Dang, X.~Liu, and Y.~Huang, ``Optimization algorithm of
  initial orbit based on internal penalty function method,'' in \emph{2007 IEEE
  International Conference on Systems, Man and Cybernetics}.\hskip 1em plus
  0.5em minus 0.4em\relax IEEE, 2007, pp. 3202--3207.

\bibitem{korf2002new}
R.~E. Korf, ``A new algorithm for optimal bin packing,'' in \emph{Aaai/Iaai},
  2002, pp. 731--736.

\bibitem{project1}
T.~Thane, ``conference-mgmt,''
  \url{https://github.com/thanethomson/conference-mgmt}, 2017, online.

\end{thebibliography}
\bibliographystyle{IEEEtran}

\end{document}